%% file: main.tex
\documentclass[letterpaper, 10 pt, conference]{ieeeconf}  

\usepackage{times}
\usepackage{epsfig}
\usepackage{graphicx}
\usepackage{amsmath}
\usepackage{amssymb}
\usepackage[noend]{algpseudocode}
\usepackage{balance}
\usepackage{color}
\usepackage{url}
\usepackage{authblk}
\usepackage{multirow}
\usepackage{subfig}
\usepackage{cite}
\usepackage[linesnumbered,ruled]{algorithm2e}
\usepackage{todonotes}
\usepackage{pgfplots}
\usepackage{tikz}
\usepackage{url}
\usepackage{hyperref}
\usepackage{soul}
\pgfplotsset{compat=newest}
\usetikzlibrary{plotmarks}
\usetikzlibrary{arrows.meta}
\usepgfplotslibrary{patchplots}
\usepackage{grffile}
\definecolor{anotherColor}{HTML}{228B22}
\definecolor{findOptimalPartition}{HTML}{D7191C}
\definecolor{storeClusterComponent}{HTML}{FDAE61}
\definecolor{dbscan}{HTML}{ABDDA4}
\definecolor{constructCluster}{HTML}{2B83BA}
\definecolor{ICPColor}{HTML}{9966CC}

\newcommand{\eg}{\emph{e.g.},}
\newcommand{\ie}{\emph{i.e.},}
\newcommand{\etal}{\emph{et~al.}}

\usetikzlibrary{shapes,arrows,positioning}
\usetikzlibrary{calc}
\tikzstyle{block} = [draw, fill=blue!20, rectangle,
    minimum height=3em, minimum width=6em]
\tikzstyle{sum} = [draw, fill=blue!20, circle, node distance=1cm]
\tikzstyle{input} = [coordinate]
\tikzstyle{output} = [coordinate]
\tikzstyle{pinstyle} = [pin edge={to-,thin,black}]

\usetikzlibrary{calc,trees,positioning,arrows,chains,shapes.geometric,%
    decorations.pathreplacing,decorations.pathmorphing,shapes,%
    matrix,shapes.symbols}

\tikzset{
>=stealth',
  punktchain/.style={
    rectangle,
    rounded corners,
    draw=black, thick,
    text width=18em,
    minimum height=1.5em,
    text centered,
    on chain},
  line/.style={draw, thick, <-},
  element/.style={
    tape,
    top color=white,
    bottom color=blue!50!black!60!,
    minimum width=8em,
    draw=blue!40!black!90, very thick,
    text width=10em,
    minimum height=2.5em,
    text centered,
    on chain},
  every join/.style={->, thick,shorten >=1pt},
  decoration={brace},
  tuborg/.style={decorate},
  tubnode/.style={midway, right=2pt},
}

\graphicspath{ {figures/} }

\IEEEoverridecommandlockouts                              
\title{\LARGE \bf Local Descriptor for Robust Place Recognition using LiDAR Intensity}

\author{Jiadong Guo \textsuperscript{\textdagger\ \textasteriskcentered}, Paulo V K Borges \textsuperscript{\textdagger}, Chanoh Park \textsuperscript{\textdagger} and Abel Gawel \textsuperscript{\textasteriskcentered}
\thanks{\textsuperscript{\textdagger} Robotics and Autonomous Systems Group, CSIRO, Pullenvale, QLD 4069, Australia.
        {\tt\footnotesize paulo.borges@csiro.au chanoh.park@csiro.au}}
\thanks{\textsuperscript{\textasteriskcentered} Autonomous Systems Lab, ETH Z\"urich, 8092 Z\"urich, Switzerland.
        {\tt\footnotesize jiadong.guo@outlook.com gawela@ethz.ch}}%
}

\begin{document}

\maketitle
\thispagestyle{empty}
\pagestyle{empty}

\begin{abstract}
Place recognition is a challenging problem in mobile robotics, especially in unstructured environments or under viewpoint and illumination changes.
Most LiDAR-based methods rely on geometrical features to overcome such challenges, as generally scene geometry is invariant to these changes, but tend to affect camera-based solutions significantly. Compared to cameras, however, LiDARs lack the strong and descriptive appearance information that imaging can provide.

To combine the benefits of geometry and appearance, we propose coupling the conventional geometric information from the LiDAR with its calibrated intensity return. This strategy extracts extremely useful information in the form of a new descriptor design, coined ISHOT, outperforming popular state-of-art geometric-only descriptors by significant margin in our local descriptor evaluation.
To complete the framework, we furthermore develop a probabilistic keypoint voting place recognition algorithm, leveraging the new descriptor and yielding sublinear place recognition performance. The efficacy of our approach is validated in challenging global localization experiments in large-scale built-up and unstructured environments.

\end{abstract}

\section{Introduction}
Place recognition using Light Detection and Ranging (LiDAR) sensors in large unstructured environments remains a challenging research problem. While LiDAR-based methods have made great progress in man-made environments, these often suffer in natural environments. Natural features, like vegetation, can be cluttered and return noisy surface normal estimates, which many geometric description methods rely on~\cite{guoComprehensivePerformanceEvaluation2016,rusuFastPointFeature2009}. Recent developments tackle the place recognition problem with segments~\cite{dubeSegMatchSegmentBased2017a}, semantics~\cite{gawelXViewGraphBasedSemantic2017} or learned global descriptors~\cite{uyPointNetVLADDeepPoint2018b}. However, these methods often require high quality ground removal or additional segmentation by image sensors~\cite{gawelXViewGraphBasedSemantic2017}, which is not straightforward for unstructured environments. To improve place recognition performance in natural environments we suggest using intensity returns, which are readily available with most modern LiDAR sensors for robotics. Our prior work~\cite{copDELIGHTEfficientDescriptor2018} shows that collecting raw intensity values into a global descriptor and using matching descriptors to filter place candidates significantly improves the recognition quality. As intensity is inherently invariant to lighting conditions, Barfoot \etal ~\cite{barfootDarknessVisualNavigation2016} and Neira \etal ~\cite{neiraFusingRangeIntensity1999} use intensity images to localize and navigate ground vehicles, even in dark environments.

Despite these many contributions, to the authors' knowledge a more flexible localization approach using intensity-based local 3D descriptors has not yet been demonstrated.

\begin{figure}[tb]
\centering{
\includegraphics[width=.46\textwidth]{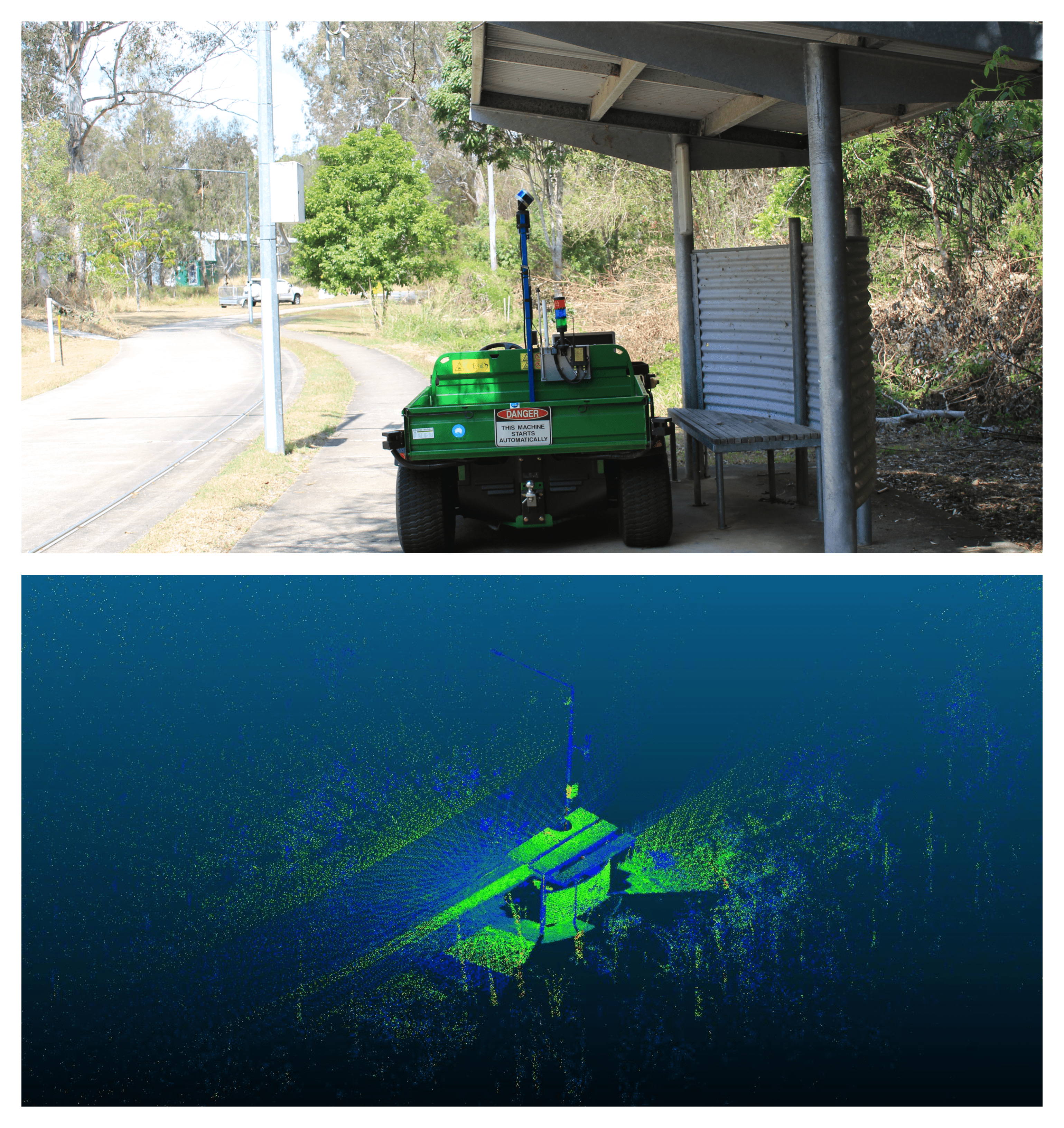}}
\caption{A challenging scenario (upper) from our \textbf{occlusion} dataset. The view from the LiDAR (mounted on top of the central pole) is heavily occluded by the nearby structure. With our proposed algorithm, the vehicle is able to recover its position in the global map using only the local LiDAR point cloud (lower image, overhead view, coloured by intensity) without prior sensor or motion information (\eg GPS, IMU, etc)}.
\label{fig:teaser}
\vspace{-7mm}
\end{figure}

In this work, we aim to overcome these obstacles by including LiDAR intensity measurements in local 3D descriptors. This descriptor, coined \textit{Intensity Signature of Histograms of OrienTations} (ISHOT), is analogous to its RGB-enriched version ColorSHOT~\cite{tombariCombinedTextureshapeDescriptor2011}. Additionally, we propose a new probabilisitic keypoint voting approach for place recognition. The algorithm is inspired by the method developed by Bosse and Zlot~\cite{bossePlaceRecognitionUsing2013}, but instead of using a voting threshold to find potential candidates, we empirically model the voting precision and update place matching probabilities for places in the database after each vote.
We evaluate our algorithm in real-world outdoor datasets, using a rotating 3D LiDAR setup, significantly outperforming classic geometry-only place recognition approaches. Our approach is able to efficiently recover the robot position even in challenging scenarios (see Fig.~\ref{fig:teaser}, where the robot, a John Deere Gator, is shown under a roof on the top image). In summary, the key contributions of this paper are:
\begin{itemize}
\item A novel intensity-enriched local 3D descriptor using calibrated LiDAR intensity return
\item An adapted keypoint voting regime, based on empirical modelling of voting precision
\end{itemize}

The method is evaluated in large-scale outdoor experiments\footnote{We make the datasets available under \href{https://doi.org/10.25919/5bff3be8c0d24}{https://doi.org/10.25919/5bff3be8c0d24}}, spanning 160,000\,$m^{2}$
This paper is organized as follows: in Section~\ref{sec:related_work} we first review related work on place recognition and LiDAR intensity. Implementation details of our descriptor ISHOT are provided in Section~\ref{sec:intensity}, while our probabilistic keypoint voting place recognition pipeline is described in Section~\ref{sec:voting}. Section~\ref{sec:experiments} presents evaluations using real-world datasets followed by a discussion of our results in Section~\ref{sec:discussion}.

\section{Related Work and Background}
\label{sec:related_work}

The classic approach towards recognizing places using 3D data is the detection, extraction, and matching of local 3D descriptors against a database of places represented by descriptors. The detection can be performed using keypoints~\cite{bossePlaceRecognitionUsing2013}, segments~\cite{dubeSegMatchSegmentBased2017a}, or complete point clouds~\cite{copDELIGHTEfficientDescriptor2018}. The surrounding neighborhood of each detected keypoint is further described using local 3D descriptor~\cite{bossePlaceRecognitionUsing2013, cieslewskiPointCloudDescriptors2016c, tombariUniqueSignaturesHistograms2010}. Next, these descriptors jointly suggest a place candidate from database using methods such as: voting~\cite{bossePlaceRecognitionUsing2013}, bag-of-words~\cite{stederPlaceRecognition3D2011} or classification~\cite{dubeSegMatchSegmentBased2017a}. Finally, a verification step confirms geometric consistency of the place recognition~\cite{aldomaTutorialPointCloud2012}. In our work, we innovate on the keypoint description step using LiDAR intensities and introduce a probabilistic keypoint voting mechanism for matching.

LiDAR sensors return the received energy level and the range for every measurement. While the range has very high resolution on some sensors (\eg \ $\pm 3\ cm$ on the VLP-16), no standardized metric exists for the intensity readings across different sensors. As a result, the intensity return is typically discarded for localization~\cite{collierRealTimeLidarBasedPlace2012} and only geometric information is kept for further processing. In an effort to calibrate the intensity return of LiDAR sensors, Levinson and Thrun~\cite{levinsonUnsupervisedCalibrationMultibeam2014} calibrate a Velodyne HD-64E S2 by deriving a Bayesian generative model of each beam's response to surfaces of varying reflectivity. Steder \etal \cite{stederMaximumLikelihoodRemission2015} solve the maximum likelihood problem by finding scaling factors in a lookup table dependent on incidence angle and measured distance for multiple LiDAR sensors from different manufacturers. The authors report impressive visual results, but do not report applying calibrated intensity values in practical tasks.

In recent years the use of intensity returns for LiDAR-based localization and place recognition has received some attention.
However, instead of using the intensity values directly, most works utilize high-level visual features extracted from intensity images~\cite{barfootDarknessVisualNavigation2016, hataRoadMarkingDetection2014, castorenaGroundEdgeBasedLIDARLocalization2018,pmlr-v87-barsan18a} to perform localization or visual odometry tasks. Intensity images have been successfully used for visual odometry and localization in dark environments using the SURF feature detector~\cite{barfootDarknessVisualNavigation2016}. The extracted edges can also be used in vehicle localization \cite{castorenaGroundEdgeBasedLIDARLocalization2018}. Cop \etal \cite{copDELIGHTEfficientDescriptor2018} presented \textit{DEscriptor of
LiDAR Intensities as a Group of HisTograms} (DELIGHT), a global point cloud descriptor, which is created from multiple histograms of raw intensity values.  By performing a descriptor matching of DELIGHT, the algorithm eliminates unlikely place candidates before proceeding to precise localization. However, using global point clouds can affect robustness. Khan \cite{khan3DRoboticMapping} utilizes calibrated intensity return of a single-beam Hokuyo sensor to improve the performance of 2D Hector SLAM~\cite{Kohlbrecher2013HectorOS}. Very recently, Barsan \etal \cite{pmlr-v87-barsan18a} propose to learn an calibration-agnostic embedding for both LiDAR intensity map and sweeps for a real-time localization approach.

Voting-based place recognition systems, such as our method, directly search the nearest neighbors of query local descriptors to identify potential matches to the database. Bosse and Zlot~\cite{bossePlaceRecognitionUsing2013} proposed a keypoint voting strategy that achieves sublinear matching performance using a novel Gestalt3D keypoint descriptor. By modeling the non-matching votes as a log normal distribution, an empirical threshold can be set to eliminate false positives among the place candidates and terminate the keypoint match. Despite scoring excellent results in loop closure for large unstructured environments, the matching is time consuming due to the large database of keypoints, and this approach can not deal with varying keypoint densities. Lynen \etal~\cite{lynenGetOutMy2015} achieves real-time localization performance on embedded system by using inverted multi-index descriptor matching strategy and covisibility filtering technique to reject outliers, despite the low ratio of true positives. We propose to use intensity-augmented 3D description and model the matching accuracy of a vote in a known environment instead, as this yields the voting process to be terminated much faster.

\section{Intensity-augmented 3D descriptor}
\label{sec:intensity}
In this section, we describe our intensity preprocessing and calibration procedure for a LiDAR sensor VLP-16, and introduce ISHOT, a novel intensity-augmented 3D descriptor.

\subsection{Intensity calibration and pre-processing}
\begin{figure}[tb]
\centering
\includegraphics[width=.48\textwidth]{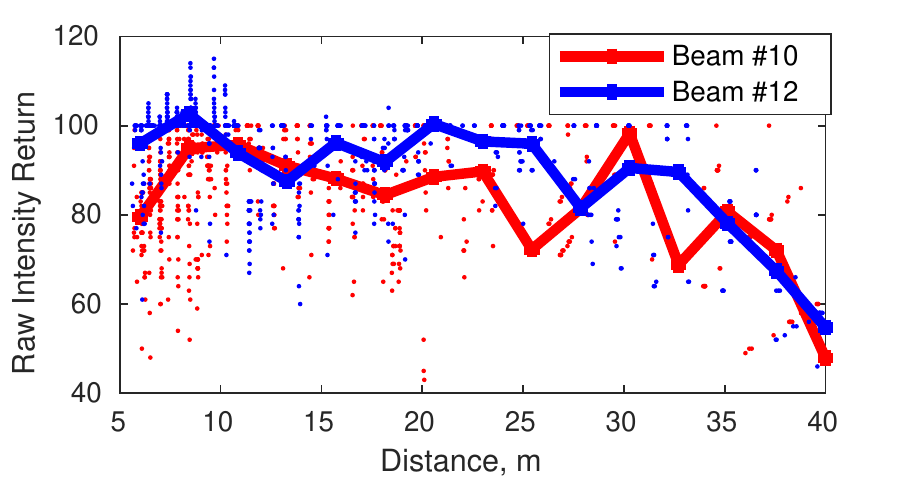}
\caption{Analysis of raw 8-bit intensity return of 2 beams on the VLP-16, pointing to a uniform high reflectance surface at incidence angle close to zero at different distances. Solid lines are fitted average from raw data. The measurements are noisy, different across beams, and degenerate at large distances. }
\label{fig:velodyne1}
\vspace{-7mm}
\end{figure}

 The Velodyne VLP-16 sensor returns a single 8-bit intensity value ($0-255$) for each measurement, corresponding to the surface's physical reflectance\footnote{This data is available since the the VLP-16 2.0 firmware update~\cite{wolfgangjuchmannVelodyneSoftwareVersion2012}.}. Intensity returns of value less than $100$ matches diffusive objects, while values above 100 are retro-reflective, \ie traffic signs. Internally, the sensor's balancing function compensates for the squared energy loss by its travelled distance, in order to return consistent results for the same surface. Although the VLP-16 can measure distances up to 100 m, as with most Lidars its intensity return degenerates at high range due to low energy return and varying resolution (Fig. \ref{fig:velodyne1}).

Therefore, we discard measurements beyond a distance threshold at $30 \ m$, and calibrate the remaining using an unsupervised Bayesian approach proposed by Levinson and Thrun~\cite{levinsonUnsupervisedCalibrationMultibeam2014}, as it does not require incidence angle computation and deals with discrete intensity returns directly. The calibration method leads to a mapping function $g_l$ for every beam $l$ between a discrete measurement $I_{measured}$ and the most likely true intensity of the surface $I_{true}$:
\begin{equation}
I_{true} = g_l(I_{measured})
\label{eq:I_lookup2}
\end{equation}

Finally, we rescale the intensity value to $[0,1]$, where an original value of 100 and beyond is mapped to 1. This is due to very unlikely encounters of retro-reflective objects in the wild and to avoid further firmware correction from the sensor.

\subsection{Constructing ISHOT}
Mathematically, a multi-cue keypoint descriptor $D$, such as ColorSHOT~\cite{tombariCombinedTextureshapeDescriptor2011}, is a chain of generalized \textit{Signatures of Histograms} $SH^i_{(G,f)}(P)$ for the support region around feature point $P$. $G$ is a vector-valued point-wise property of a vertex, and $f$, the metric used to compare two of such point-wise properties.
\begin{equation}
D(P)= \bigcup ^m _ {i=1}SH^i_{(G,f)}(P)
\label{eq:2}
\end{equation}
Here, $m$ denotes the number of different data cues.
Inspired by ColorSHOT~\cite{tombariCombinedTextureshapeDescriptor2011}, we use two types of cues $m=2$, consisting of a geometric component \textit{Signature of Histograms of OrienTations} SHOT~\cite{tombariUniqueSignaturesHistograms2010} and a texture component using calibrated intensity returns. Our matching metric $f$ is the difference between each sample inside the support region $Q$ and the feature point $P$:

\begin{equation}
f(I_P, I_Q)= I_P - I_Q
\label{eq:3}
\end{equation}

The histogram of intensity differences is configured to have 31 bins in each of the 32 spatial support regions inside a keypoint's neighborhood. The definition of the support regions are identical to the original formulation from SHOT~\cite{tombariUniqueSignaturesHistograms2010}. Together with the 352 dimensions of the original SHOT descriptor, ISHOT has 1344 feature dimensions in our configuration.
\pgfdeclarelayer{background}
\pgfdeclarelayer{foreground}
\pgfsetlayers{background,main,foreground}
\begin{figure*}[bt]
\centering
\begin{tikzpicture}[block/.style   ={rectangle, draw, text width=4.0em, text centered, rounded corners, minimum height=2em, minimum height=6.1em}, blocki/.style   ={rectangle, draw, text width=4.0em,text centered, rounded corners, minimum height=3em, minimum height=1.5em}, blockj/.style   ={rectangle, draw, text width=4.0em, text centered, rounded corners, minimum height=3em, minimum height=2.5em},
node distance=3.0cm, auto, >=latex']

    \node [align=center] (gator) {\includegraphics[width=9.0em]{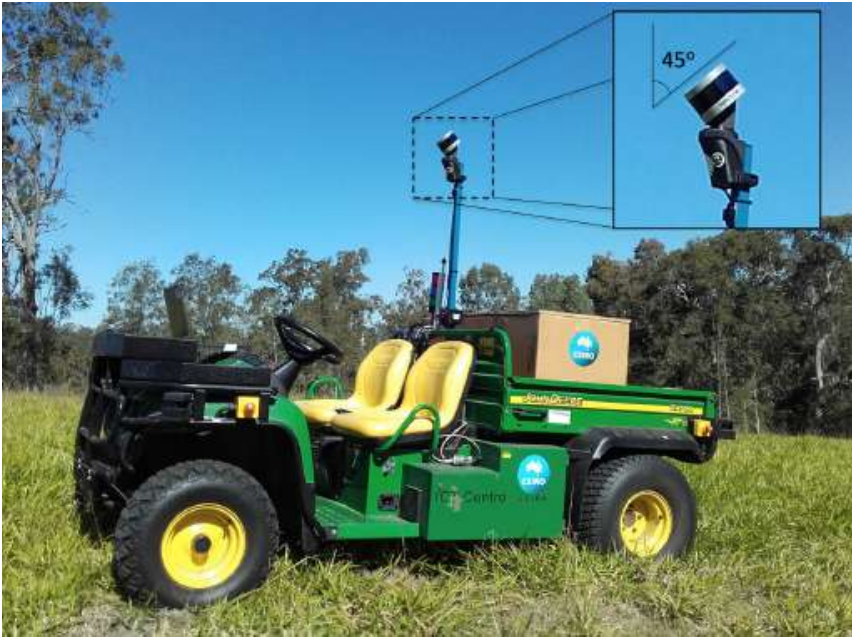}\\ \scriptsize The robot and its sensor};
	\node [align=center, right of = gator] (data) {\includegraphics[width=7.0em]{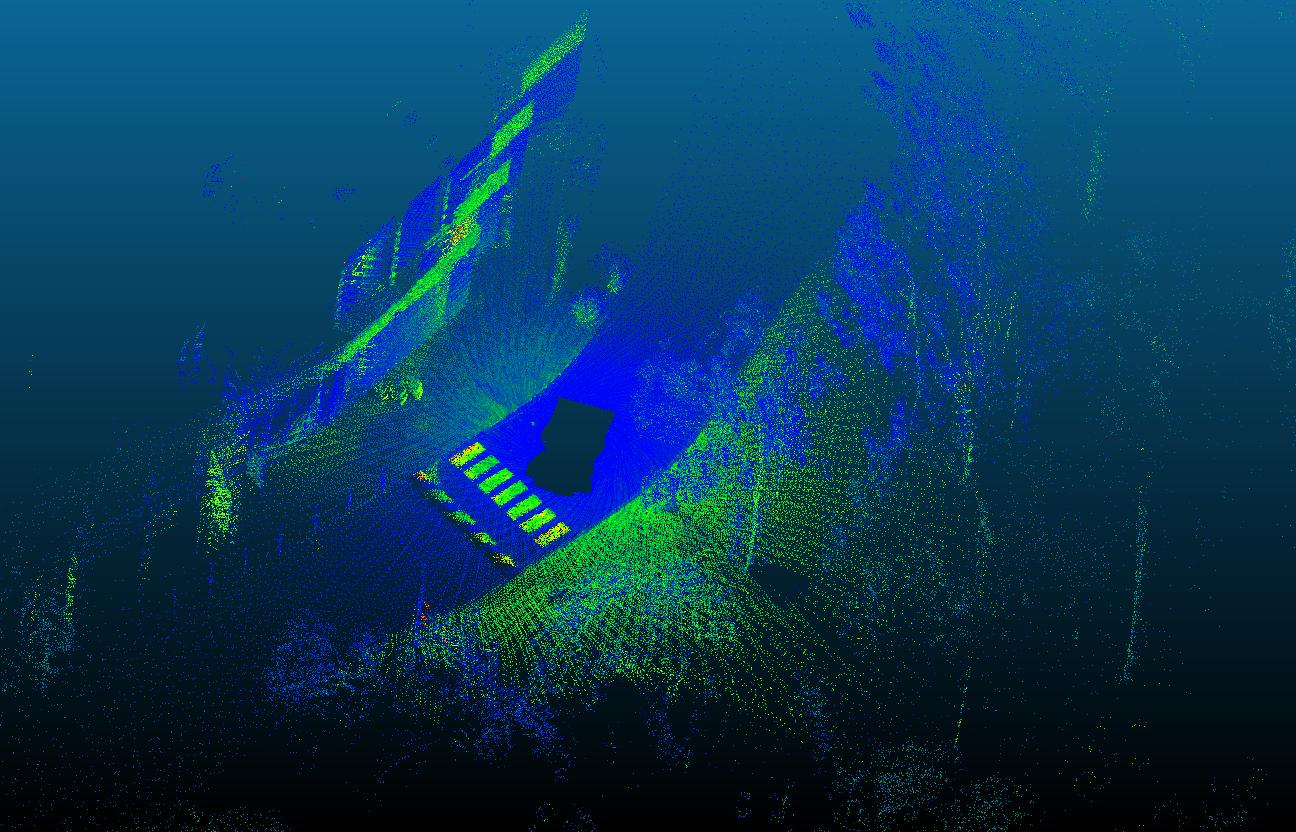}\\ \scriptsize 3D LiDAR Scan};
	\node [blocki, align=center, right of = data] (abstraction) {
	\scriptsize ISHOT Feature \\ \scriptsize extraction};
	\node [blocki, align=center, right of=abstraction] (matching) {
	\scriptsize Probabilistic \\ \scriptsize place voting};
	\node [blocki, align=center, right of=matching] (estimation) {
	\scriptsize SHOT Local \\ \scriptsize refinement };
	\node [align=center, right of = estimation] (output) {\includegraphics[width=7.0em]{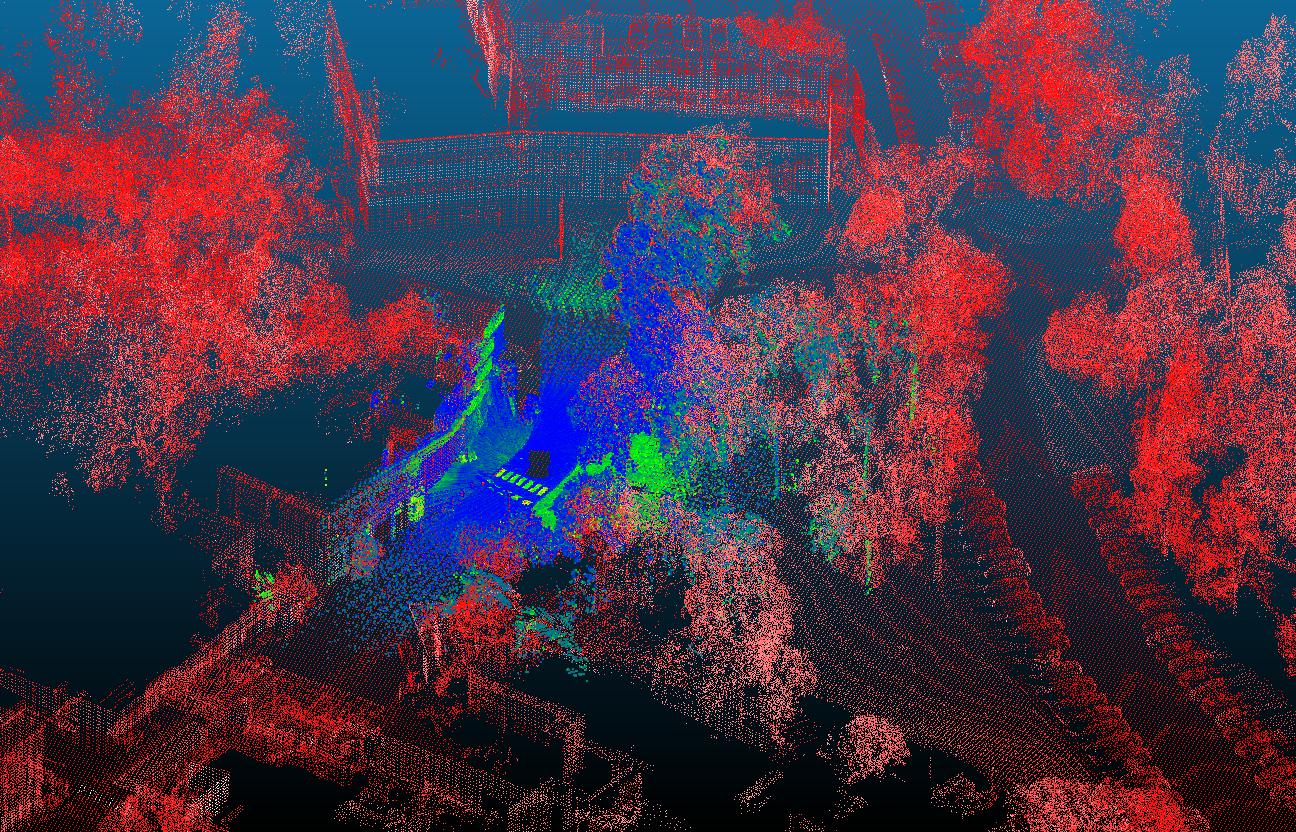} \\ \scriptsize Global Localization};

\node [blockj, align=center, below = 0.4 of abstraction] (global) {
 \scriptsize Global map \\ \scriptsize of places};

\draw[->] (data.east) -- (data-|abstraction.west);
\draw [->] (abstraction) -- node {} (matching);
\draw [->] (matching) -- node {} (estimation);
\draw [->] (estimation) -- node {} (output);
\draw [<-] (matching.north west)+(-2.4,0) arc (180:0:1.4cm and 0.4cm);
\draw [->] (global) -| node {} (matching.south);
\draw [->] (global) -| node {} (estimation.south);
\draw [<-] (estimation.south west)+(-5.2,0) arc (-180:0:2.75cm and 0.4cm);;
\begin{pgfonlayer}{background}
\path (abstraction.north west)+(-0.2,0.5) node (a) {};
\path (estimation.south -| estimation.east)+(+0.2,-0.5) node (b) {};
\path[rounded corners, draw=red, dashed]
    (a) rectangle (b);
\end{pgfonlayer}
\end{tikzpicture}
\caption{Intensity augmented 3D place recognition system overview. A scan is generated by two full rotation of the tilted LiDAR sensor. Keypoints are first detected by ISS-BR in current 3D scan. A batch of ISHOT descriptors are extracted to be matched against the global database, voting for the most likely place candidate. Once match probability passes a threshold, the system proceeds to the geometric consistency refinement stage between scan and database candidates.}
\label{fig:overview}
\vspace{-2mm}
\end{figure*}
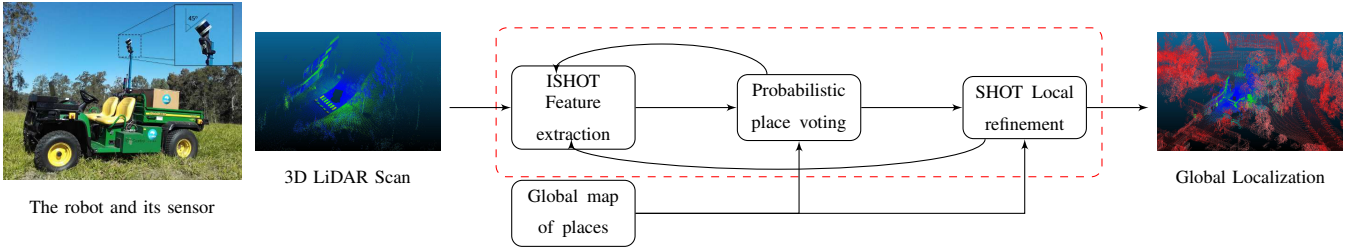
\section{Probabilistic keypoint voting }
\label{sec:voting}
We developed our approach based on the keypoint voting pipeline from Bosse and Zlot~\cite{bossePlaceRecognitionUsing2013}; however employing our ISHOT 3D local descriptors with the \textit{Intrinsic Signature Shapes with Boundary Removal} (ISS-BR)~\cite{zhongIntrinsicShapeSignatures2009} keypoint detection. This keypoint sampling technique leads to better matching performance~\cite{guoComprehensivePerformanceEvaluation2016} but results into environment-specific keypoint densities. To address this uneven keypoint distribution and leverage the efficacy of ISHOT, we further propose a probabilistic voting approach that updates probabilities of correct place matches by modeling the closest neighbor match voting accuracy in a known environment.

\subsection{Place Recognition Pipeline Overview}

Our place recognition system (Fig.~\ref{fig:overview}) is based on descriptor matching and place voting. Inputs to our system are local 3D scans and a global feature map discretized into places with the aim to localize the local scan within the global map. A scan consists of all LiDAR measurements accumulated during two full rotations of the actuated sensor, while the vehicle remains stationary. Every individual point is timestamped and projected into the vehicle's frame based on the motor encoder's reading at the given timestamp. We first extract ISHOT descriptors from the calibrated local 3D LiDAR scan and then match them against descriptors from places of the global map, voting for the most probable place. After narrowing the search to candidate places within the global map, we perform 3D feature matching between the scan and the candidate places to refine our estimation. The resulting candidate matches are registered using \textit{Iterative Closest Point} (ICP) to obtain the final transformation between robot and map.
\subsection{Global Places Database \& Localization Query}
\label{sec:database}
A global map is partitioned into discrete places along the trajectory associated (in a SLAM sense) to its creation. The centers of the places are set a minimum distance apart from each other and consist of all measurements within a time window. This distance is set much less then the range of LiDAR detection, so that nearby places overlap. This point cloud is down-sampled by voxelization and the intensity values are corrected with a mapping function and averaged over each voxel. The ISS-BR detector is then used to detect keypoints on the downsampled point cloud of each place. In the last step all keypoints are described using ISHOT, serialized and saved in a database of global places for later retrieval.

When localization is requested, the robot captures a local 3D LiDAR scan of the environment. The local point cloud is then processed similarly to the places. Each resulting feature descriptor from the local point cloud is then matched against the global database of descriptors from all places.

\begin{figure*}[bt]
\centering{
\def\svgwidth{70mm}
\includegraphics[height=.24\textheight,trim={2.6cm 4.48cm 2.0cm 2.0cm},clip]{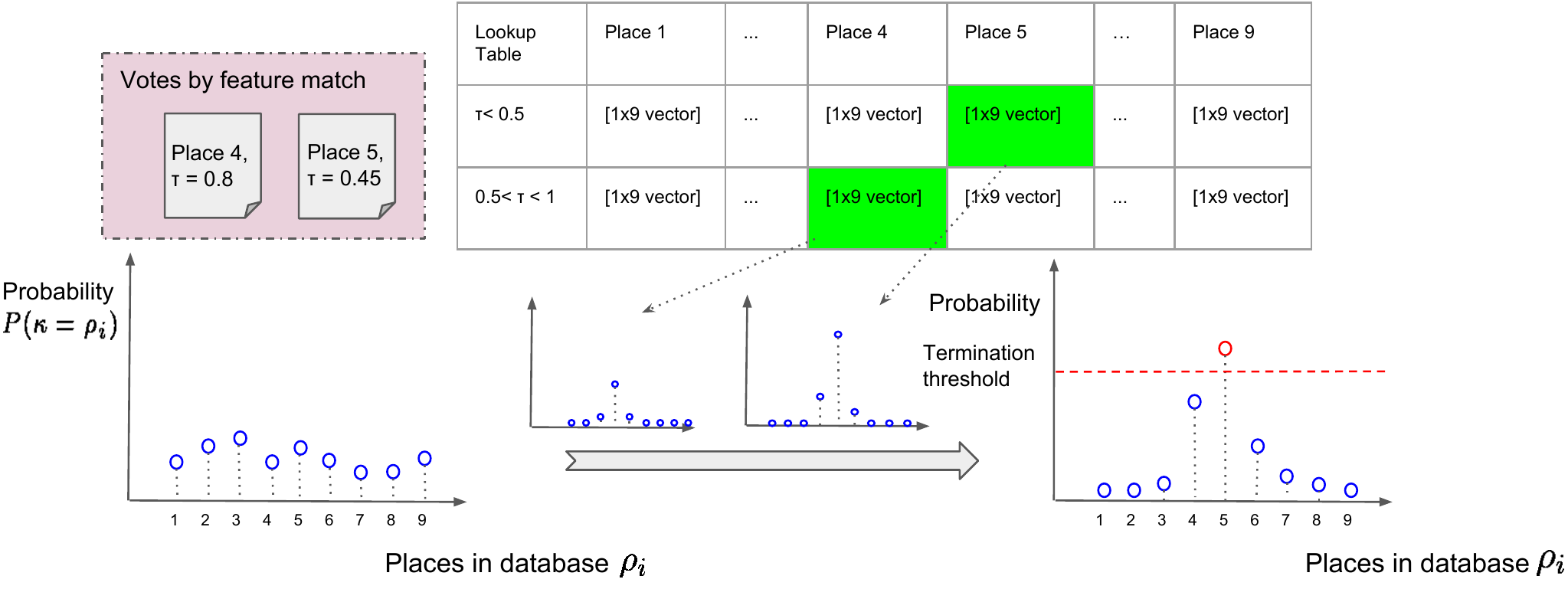}
}
\caption{Probability update during a voting process. This example depicts 2 votes from 9 places and 2 $\tau$ ranges. For every vote, a precomputed probability is extracted from the lookup table and merged into the matching probability of places from the database. Instead of voting for one place specifically, a vote updates the probability distribution of all places.}
\label{fig:update}       
\vspace{-3mm}
\end{figure*}
\subsection{Probabilistic voting}
Our probabilistic voting process considers only the two nearest neighbors for every matched descriptor. The place where the nearest neighbor in the database is extracted from counts as a place candidate $\rho_v$, while the \textit{Nearest Neighbor Distance Ratio} (NNDR) to the second closest match is recorded as a quality measure, similar to the measure presented by Lowe~\cite{Lowe2004}. This NNDR quality measure $\tau$ is defined as follows:
\begin{equation}
\tau = \frac{||\mathbf{d_q}-\mathbf{d_b}||}{||\mathbf{d_q}-\mathbf{d_b'}||}
\label{eq:8}
\end{equation}
where $\mathbf{d_q} \in {\mathbf{R}^{1344}}$ denotes the query ISHOT descriptor in scan, $\mathbf{d_b}$ and $\mathbf{d_b'}$ are the closest and second closest neighboring descriptor in the database, respectively. We now define a vote $v$ to consist of a place candidate $\rho_v$ and its quality measurement $\tau_v$:
\begin{equation}
v = \{\rho_v, \tau_v\}
\end{equation}
Assuming each vote is independent, we can update the probability that the current \textit{scan} $\kappa$ is matched to a place in the database $P(\kappa = \rho_i)$ given $q$ votes:
\begin{equation}
P(\kappa = \rho_i) = \eta \cdot \prod_{m=1}^q P(\rho_i|v_m)
\label{eq:9}
\end{equation}
where $\eta$ is a normalization factor and the probability $P(\rho_i|v_m)$ is precomputed at different $\tau$ and any place $\rho_i$ for a given descriptor by modeling the voting precision.

\subsection{Modeling voting precision}

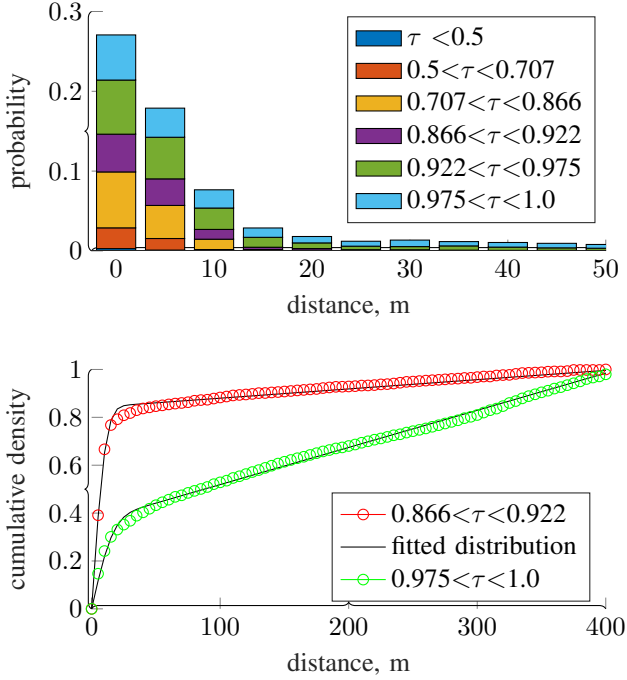
\begin{figure}[bt]
\centering{
\def\svgwidth{70mm}
\input{2Plots.tex}

\caption{(upper) Distribution of votes resembles half-normal distribution with a uniformly distributed long tail. The long flat tail up to 400 m is omitted for visibility.  Lower $\tau $ ranges correspond to higher matching quality, and appear less frequent. (lower) Fitting parameters for two ranges of $\tau$ using the cumulative distribution function.}
\label{fig:cdf2}       
\vspace{-7mm}
}
\end{figure}

We model our voting process as a mixed probability distribution, formed from a half-normal and a uniform distribution, dependent on the distance to its real location and matching score $\tau$. The voting process can be modeled as a normal distribution, where places spatially close to ground truth locations are most likely to receive the vote~\cite{gehrigVisualPlaceRecognition2016},~\cite{bossePlaceRecognitionUsing2013}. As we use distances to model this likelihood, we fold the normal distribution to become a half normal distribution with zero mean. The additional uniform distribution accounts for the probability of finding random non-matches and gives the distribution a long tail, as in Fig. \ref{fig:cdf2}. Collectively, they form the matching probability of place $\rho_i$ given a vote:

\begin{equation}
P(\rho_i|d(\rho_v,\rho_{gt})) = \lambda(\frac{\sqrt{2} }{\sigma \sqrt{\pi}} \exp(\frac{-d(\rho_v,\rho_{gt})^2}{2\sigma^2})) + (1-\lambda)
\label{eq:dist_fit}
\end{equation}

The probability of matching a place $P(\rho_i)$ is thus dependent on the place's distance to ground truth $d(\rho_v,\rho_{gt})$, where $\lambda$ balances the ratio between the two probabilities and $\sigma$ is the variance for the normal distribution. For a given descriptor and a range of $\tau$, the two parameters $\sigma, \lambda$ are found by fitting the theoretical curve of a \textit{Cumulative Distribution Function} (CDF) to training data matched using ground truth, see Fig.~\ref{fig:cdf2}.

\subsection{Probability update and terminate condition}
To avoid the expensive descriptor computation and matching in a high dimensional space, we leverage the improved quality of our descriptor by only computing and matching them when needed. In every iteration, we compute and match features for a randomly selected subset of unprocessed keypoints, and update the matching probability of all places in the database from the votes. Given the fitted distribution of voting for each $\tau$ range, we precompute a matching probability for every place at every possible vote pair ${\rho_v, \tau_v}$ by approximating the ground truth $\rho_{gt}$ with the current voted place $\rho_v$ using (\hspace{1sp}\ref{eq:dist_fit}).

This probability is computed for every place in the database and normalized to sum to 1. It can be interpreted as a confidence metric that determines whether the current vote is coming from a place $i$, given the matching quality $\tau$ and the spatial relationship of places in the database. After the feature matching, for every vote $\{\rho_v, \tau_v\}$ in the batch, a vector is extracted from the table to update the matching probability of place $i$ in the database, see Fig.~\ref{fig:update}.
We apply additional normalization to account for the different keypoint densities within each place in the database. If any candidate has a probability that surpasses a certain threshold $\xi_v$, the algorithm proceeds to the geometric verification step directly, skipping the computation and matching of further features. If the given threshold is never reached after all keypoint descriptors are matched, the places are checked by their voting scores.

\subsection{Fine registration and geometric consistency check}
Once the probability of a place surpasses the acceptance threshold, we roughly align the current scan against the candidate place by matching the local features with features from the place candidate and finding geometric consistent transformations. The matching is simpler as the database only consists of keypoints from one place, but it needs to be run multiple times for multiple candidate places. Here, SHOT is chosen for its much faster matching while preserving relatively high quality feature matching. Starting from this initial estimate, we apply point-to-plane ICP between the voxelized candidate place and current scan for the fine registration, and accept the registration result if the remaining sum of squared distance does not surpass an empirically determined threshold $\epsilon_{ICP}$.

\section{Experiments}
\label{sec:experiments}

We now evaluate our approach on several real-world datasets. We first introduce the datasets, then present experimental findings for isolated and integrated experiments. We benchmark the proposed ISHOT descriptor against popular geometric descriptors in an \textit{Area Under the precision-recall Curve}
(AUC) evaluation similar to Guo \etal~\cite{guoComprehensivePerformanceEvaluation2016}. Finally, we compare the full probabilistic voting pipeline using ISHOT against reference localization approaches.

\subsection{Datasets}
The benchmarking dataset consists of one large map and three sets of LiDAR \textit{scan}s. The datasets were generated outdoors at \textit{Queensland Centre for Advanced Technologies}  (QCAT) in Brisbane, Australia.

\begin{figure}[bt]
\centering{
\def\svgwidth{70mm}
\includegraphics[height=.24\textheight]{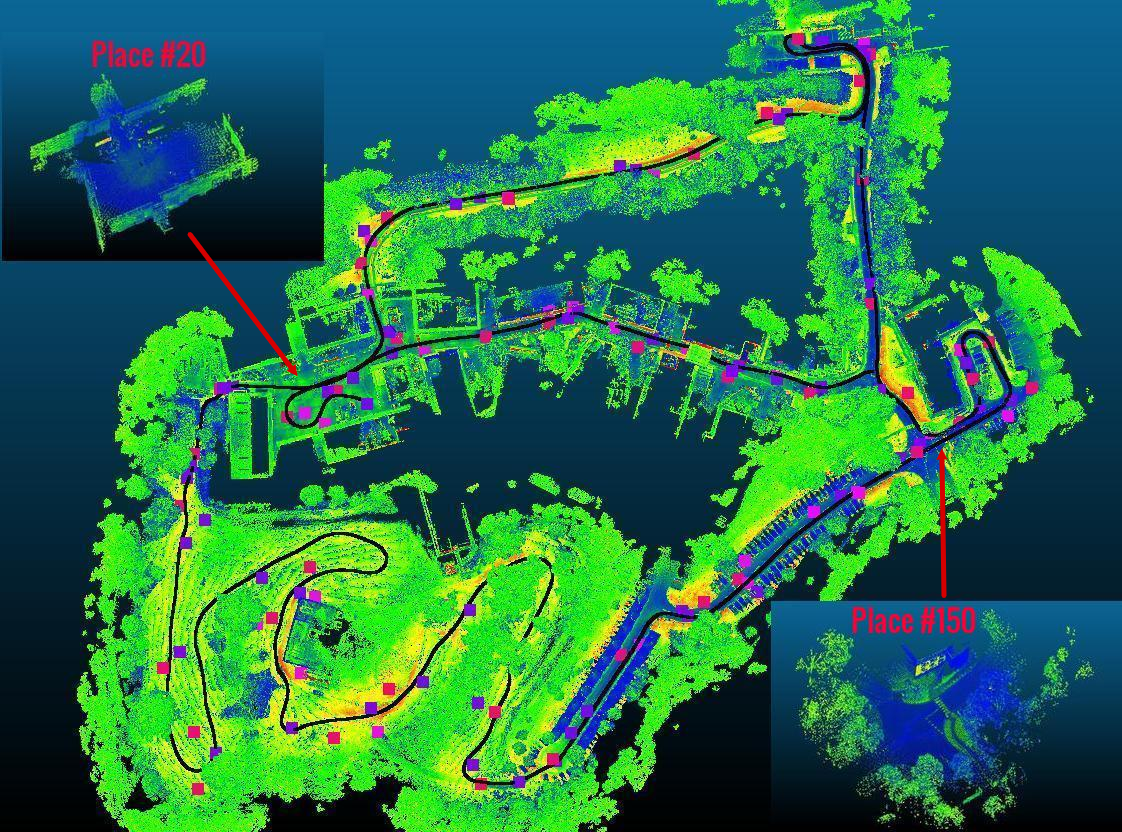}
}
\caption{Map of QCAT. Pre-processed map with colors correspond pre-processed intensity value. Black curve is the trajectory during map creation. Squared markers are the ground truth location of where the static scans took place. purple:  \textbf{gator} dataset, pink: \textbf{pole} dataset, magenta: \textbf{occlusion} dataset. Two example places are shown in the corner.}
\label{fig:dataset_01}       

\end{figure}

The environment was first mapped with a state-of-the-art SLAM algorithm~\cite{bosseContinuous3DScanmatching2009}
using our autonomous ``Gator" platform~\cite{romeroEnvironmentawareSensorFusion2016}, which is equipped with a rotating 3D LiDAR sensor, as shown in Fig.~\ref{fig:overview}. From the point cloud we generate 438 places, covering an area of approximately 160,000 $m^2$. The three sets of scans were generated using the same sensor setup under different conditions from ``easy" to ``hard". All scans were collected within $10\ m$ from the original trajectory(see Fig.~\ref{fig:dataset_01}), and processed to include two rotations of encoder.

\begin{enumerate}
\item \textbf{gator} dataset: consists of 58 static scans, generated by the mapping vehicle on diverse locations on the map.
\item \textbf{pole} dataset: consists of 41 static scans using the same sensor module on the pole independent from the mobile platform. The pole is held at diverse heights between $0.8\ m$ to $2.5\ m$ around the site to create viewpoint differences. The point clouds are gravity aligned.
\item \textbf{occlusion} dataset: consists of 31 static scans generated by the mobile platform over the complete map. The field of view is partially occluded by buildings, cars, passengers and industrial items.
\end{enumerate}

Ground truth transformations for all datasets with respect to the global map is obtained by manually aligning the point clouds. Additionally, we record a calibration dataset for the intensity calibration of the LiDAR sensor which consists of a 120 seconds driving sequence with the Gator platform on the QCAT site.

\subsection{Evaluation of local descriptors}
\begin{figure}[tb]
\centering{
\def\svgwidth{70mm}
\includegraphics[height=.2\textheight]{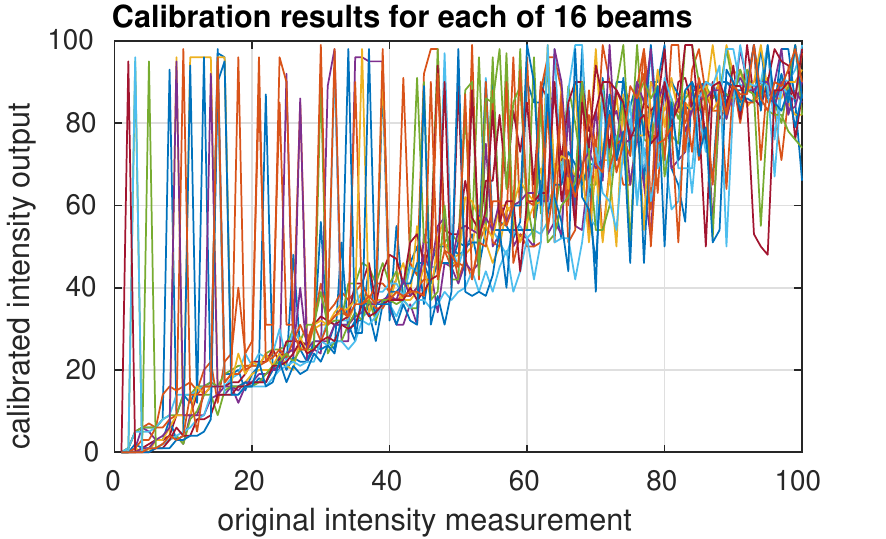}
}
\caption{The expected environment intensity given each beam’s intensity
return between 0 and 100. All 16 beams are shown here, note there is significant variation between beams.}
\label{fig:16beams}       
\vspace{-3mm}
\end{figure}

Firstly, we calibrate the LiDAR sensor according to Section~\ref{sec:intensity} on the calibration dataset. Fig.~\ref{fig:16beams} shows the calibration result of all 16 beams on VLP-16.

For evaluating the descriptive power of ISHOT, we use the \textbf{gator} and \textbf{pole} datasets, where individual descriptors are matched against their nearest neighbors in the map descriptor database. We use precision, recall, and AUC as performance measures for this evaluation. True positives are counted if a matched descriptor originates from a place that falls within $5\ m$ of the ground truth location. The results are compared against a range of popular 3D descriptors, including \textit{Unique Shape Context} (USC)~\cite{tombariUniqueShapeContext2010}, \textit{Fast Point Feature Histograms} (FPFH)\cite{rusuFastPointFeature2009}, \textit{Neighbour-Binary Landmark density Descriptor} (NBLD)~\cite{cieslewskiPointCloudDescriptors2016c}, Gestalt3D~\cite{bossePlaceRecognitionUsing2013} and SHOT~\cite{tombariUniqueSignaturesHistograms2010};
 see Table~\ref{tbl:descriptor}. We vary the $\tau$ ratios as defined in (\hspace{1sp}\ref{eq:8}) from $0.5$ to $1$ for generation of the performance measures.

\begin{table}[tb]
\centering
\caption{Local 3D descriptors used in the evaluation. }
\begin{tabular}{cccc}
\hline\noalign{\smallskip}
 descriptor & summary & size                               \\
\hline
\noalign{\smallskip}
USC & histogram of point distribution & 1980 \\
FPFH & histogram of geometric features & 33 \\
Gestalt3D & signature of point distribution & 130  \\
NBLD & binary signature of point distribution & 1408 \\
SHOT & signature of histograms of orientations & 352 \\
ISHOT &  \begin{tabular}{@{}c@{}}signature of histograms of \\ orientations and intensity\end{tabular} & 1344 \\
\noalign{\smallskip}\hline\noalign{\smallskip}

\end{tabular}

\label{tbl:descriptor}
\vspace{-2mm}
\end{table}

The ISS-BR keypoint detector salient radius is set to $2.4\,m$.
To ensure a fair comparison, we select parameters such that each local descriptor describes a similar volume of the point cloud. For all descriptors, we choose $7\,m$ for the radius and $12\,m$ as the height for structural descriptors (NBLD and Gestalt3D). Furthermore, we discard all measurements over $40\,m$ to ensure sufficient point density and downsample the point clouds using a voxel grid of $0.4\,m$, as recommended by previous work~\cite{bossePlaceRecognitionUsing2013}. The average is taken for all intensity measurement inside the voxel. For every descriptor, a database of 154,800 features are generated from the 438 places of the map, against which we match the grand total of 29,342 features extracted from 99 scans, meaning 296 keypoints per scan on average.

\begin{figure}[tb]
\subfloat{\input{auc_gator.tex}} \vspace{-2mm}\\
\subfloat{\input{auc_pole.tex}}
\caption{Local descriptor evaluation results on \textbf{gator} dataset (upper) and the more difficult \textbf{pole} (lower) datasets. }

\vspace{-3mm}
\label{fig:auc_iss}       
\end{figure}

The experimental results are depicted in Fig.~\ref{fig:auc_iss} and Table \ref{tbl:auc}. ISHOT uses raw intensity returns, while ISHOT-C is using calibrated intensity returns. The results illustrate that ISHOT outperforms all benchmarked descriptors significantly with ISS-BR detector, showing the ability to disambiguate places using intensity returns. The intensity calibration increases the descriptiveness further yielding an improvement of 73.85\% and 59.35\% against the best geometric only descriptor SHOT. Interestingly, ISHOT-C improves the overall recall rate by giving up some precision at low $\tau$, compared to ISHOT. By mapping distinctive measurements to fewer statistically dominant ``true'' values, calibration indeed helped bringing similar surfaces closer in the descriptor space. Table \ref{tbl:auc} also shows averaged processing times for feature description and matching for each scan\footnote{This result is using floating point matching. Leveraging the binary nature of the descriptor, the matching time of NBLD shall reduce significantly.}. This benchmark is timed on Intel i7-4910MQ CPU using Open-source library libNabo~\cite{elsebergComparisonNearestneigboursearchStrategies2012} for descriptor matching.

\begin{table}[t]
\caption{AUC and matching time of different descriptors.}
\centering
\begin{tabular}{cccccc}
\hline\noalign{\smallskip}
 descriptor & \multicolumn{2}{c}{AUC score}  &  \multicolumn{2}{c}{rel. AUC } & averaged  \\
  & \textbf{gator} & \textbf{pole} & \textbf{gator} & \textbf{pole} & time(s)\\
\hline
\noalign{\smallskip}
USC & 0.0071 & 0.0059 &-91.64\%& -89.52\% & 27.477 \\
FPFH & 0.0117 & 0.0135 &-86.23\%& -76.71\% & 2.221 \\
Gestalt3D & 0.0257 & 0.0063 &-69.74\%& -89.13\% & \textbf{1.702} \\
NBLD & 0.0545 & 0.0173 &-35.80\% & -69.99\% &9.522 \\
SHOT & 0.0849 & 0.0578 &0\% & 0\% &2.855 \\
ISHOT & 0.1454 & 0.0883 &+71.12\%& +52.85\% & 7.570 \\
ISHOT-C & \textbf{0.1477}& \textbf{0.0921}& +73.85\%& +59.35\% & 7.311 \\
\noalign{\smallskip}\hline\noalign{\smallskip}
\end{tabular}

\label{tbl:auc}
\vspace{-5mm}
\end{table}

\subsection{Evaluation of place recognition}
We evaluate the full probabilistic place recognition algorithm on the \textbf{gator}, \textbf{pole} and \textbf{occlusion} datasets. As performance metric, we measure the success rate of the localization, i.e., the ratio of providing a pose estimation in the global map within 3 meters of manually labeled ground truth location. For comparison, we benchmark our approach against several global localization methods, i.e., global geometric-based registration using SHOT similar to Rusu \etal~\cite{rusuFastPointFeature2009}, the original keypoint voting approach with Gestalt3D features~\cite{bossePlaceRecognitionUsing2013}, and the DELIGHT place recognition approach~\cite{copDELIGHTEfficientDescriptor2018}. In global geometric-based registration, keypoints from local scans are matched against keypoints from global map without discretization into separate places. The latter two approaches are introduced in the previous Section~\ref{sec:related_work}.
\
\begin{table*}[t]
\caption{ Evaluation of Different Global Localization Pipelines }
\centering
\begin{tabular}{lcccccccc}
\hline\noalign{\smallskip}
Pipeline  & \multicolumn{2}{c}{(a) SHOT global}  &  \multicolumn{2}{c}{(b) Gestalt Keypoint Voting } & \multicolumn{2}{c}{(c) DELIGHT 2-stage wake-up}  & \multicolumn{2}{c}{(d) ISHOT probabilistic voting } \\
 \cline{2-9}
Description&       success rate & time(s)&  success rate & time(s) &   success rate & time(s) &  success rate & time(s)  \\
\hline
\noalign{\smallskip}
  gator & 53.4\%&  26.25(4.37)&  91.3\% & 18.47(16.86)&   98.1\% &  \textbf{5.7(2.1)}&  \textbf{100}\%&  8.06(5.45)\\
  pole &  65.9\%&  8.38(3.63)&  78.1\% &  22.35(16.57)&  80.5\% &  10.51(\textbf{2.72})& \textbf{92.7}\% &  \textbf{6.20}(3.74)\\
  occlusion &  67.7\%&  22.3(4.29)&  74.2\% &  22.39(21.48)&  80.6\% &  8.67(\textbf{2.55})& \textbf{93.5}\% &  \textbf{7.00}(3.77)\\
\noalign{\smallskip}\hline\noalign{\smallskip}
  overall &  60.8	\%&  19.7 &  83.1\% &  20.63&  88.4\% &  7.93& \textbf{96.1}\% &  \textbf{7.22}\\
  \noalign{\smallskip}\hline\noalign{\smallskip}
\end{tabular}

\label{tbl:pipeline_compare}
\vspace{-5mm}
\end{table*}

The same geometric verification module is used for all pipelines with a minimum cluster size of 8 and consensus set resolution of $7\ m$. The ICP error threshold $\epsilon_{ICP}$ is chosen at $7\ m^2$. The global geometric-based registration matches $K=4$ closest neighboring features in the database to establish keypoint correspondences. Gestalt3D keypoint voting uses a uniform sampling of 10\% and matches against $K=10$ closest features, with the voting threshold set to parameters from the original paper~\cite{bossePlaceRecognitionUsing2013}. The ISHOT probabilistic voting process has a batch size of 16.
The threshold $\xi_v$ is chosen empirically at 15\%.

By modeling the matching probability and considering the spatial relationship between places, we validate that our probabilistic voting approach can terminate the matching process much earlier than the original voting process, while scoring higher precision. In fact, if the scan originates from industrial areas within the map, the matching is typically terminated after just one batch.

Table~\ref{tbl:pipeline_compare} lists the success ratio and the average and median (in parentheses) time spent for different place recognition approaches, using the same hardware as in the previous evaluation. While the Gestalt3D keypoint voting and DELIGHT approaches achieve competitive success rates on all datasets, our approach consistently achieves success rates over 90\%, with 100\% success rate on the \textbf{gator} dataset. It outperforms the reference algorithms on the more challenging datasets by a margin of over 10\%. the processing times are similar to DELIGHT, but slightly faster on average. The median time is often much lower than the average, as few difficult scans require the pipelines to extract and match all features or examine multiple candidates, drastically increase the processing time.

\begin{figure}[tb]
\centering

\begin{tikzpicture}
\begin{axis}[
    xbar stacked,
    legend style={
    legend columns=2,
        at={(xticklabel cs:0.5)},
        anchor=north,
        draw=none
    },
    ytick=data,
    axis y line*=none,
    axis x line*=bottom,
    tick label style={font=\footnotesize},
    legend style={font=\footnotesize},
    label style={font=\footnotesize},
    xtick={0,1000,2000,3000,4000},
    width=.48\textwidth,
    bar width=4mm,
    xlabel={Time in ms},
    yticklabels={time(ms)},
    xmin=0,
    xmax=4000,
    area legend,
    y=8mm,
    enlarge y limits={abs=0.625},
]
\addplot[anotherColor,fill=anotherColor] coordinates
{(340,0)};
\addplot[findOptimalPartition,fill=findOptimalPartition] coordinates
{(230,0)};
\addplot[constructCluster,fill=constructCluster] coordinates
{(2915,0)};
\addplot[dbscan,fill=dbscan] coordinates
{(280,0)};
\legend{Pre-processing, Detect \& Describe ISHOT, ISHOT Matching \& Voting , Geom. consistency check}
\coordinate (A) at (170,4mm);
\coordinate (B) at (465,4mm);
\coordinate (C) at (2000,4mm);
\coordinate (D) at (3620,4mm);
\coordinate (E) at (3640,4mm);
\coordinate (F) at (3100,4mm);
\end{axis}
\node at (A) {\footnotesize{340}};
\node at (B) {\footnotesize{230}};
\node at (C) {\footnotesize{2915}};
\node at (D) {\footnotesize{280}};
\end{tikzpicture}
\caption{Profiled performance (milliseconds) of our probabilistic keypoint voting place recognition pipeline}
\label{fig:timing_prob}
\vspace{-7mm}
\end{figure}
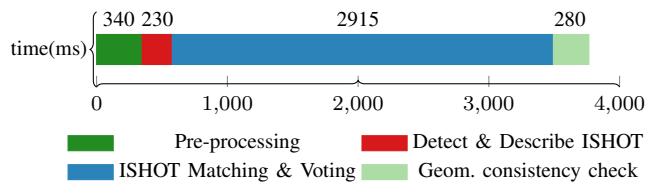

Additionally, we run the pipeline on a challenging driving dataset\footnote{\url{https://research.csiro.au/robotics/ishot/}}, and achieve an averaged global localization update rate at 0.25\,Hz on same hardware as in Table~\ref{tbl:auc}. The scene is corrupted by dynamic objects and distorted by vehicle's motion. We profiled the system's performance as in Fig.~\ref{fig:timing_prob}. The ISHOT descriptor matching in high-dimensional space is the current bottleneck of the algorithm.

\section{Discussion \& Limitations}
\label{sec:discussion}

Intensity return generates highly reproducible and informative results for a LiDAR sensor. Previous literature in robotics that involve using uncalibrated intensity returns do not report performance across different sensors. The internal pre-processing of measurements by the sensor used in our experiments is unobservable to the user, rendering the localization between different sensors difficult. Velodyne attributes such pre-processing to the need of a clear distinction between retro-reflective and diffusive objects by their customers and the device is not able to provide raw sensor energy return.

We notice a clear need for general LiDAR intensity calibration standards as our results indicate the strong benefits of intensity values in robot localization.


\section{Conclusions}

We presented ISHOT, a local descriptor combining geometric and texture information of a LiDAR sensor using calibrated LiDAR intensity returns. The descriptor outperforms state-of-art geometric descriptors by a significant margin in real world local descriptor evaluations. We furthermore propose a probabilistic keypoint voting place recognition pipeline and evaluate our work in challenging outdoor experiments. The proposed framework achieves competitive real-time global place recognition performance while being robust to viewpoint changes and occlusion. In the future we aim to find more general procedure to use intensity return across a variety of LiDAR sensors.






\bibliographystyle{IEEEtran}
\input{main.bbl}

\end{document}

%% file: 2Plots.tex
%
%
\definecolor{mycolor1}{rgb}{0.00000,0.44700,0.74100}%
\definecolor{mycolor2}{rgb}{0.85000,0.32500,0.09800}%
\definecolor{mycolor3}{rgb}{0.92900,0.69400,0.12500}%
\definecolor{mycolor4}{rgb}{0.49400,0.18400,0.55600}%
\definecolor{mycolor5}{rgb}{0.46600,0.67400,0.18800}%
\definecolor{mycolor6}{rgb}{0.30100,0.74500,0.93300}%
\begin{tikzpicture}

\begin{axis}[%
width=2.691in,
height=1.254in,
at={(0.514in,2.407in)},
scale only axis,
bar width=4,
xmin=-2.5,
xmax=50,
xlabel style={font=\color{white!15!black}},
xlabel={distance, m},
ymin=0,
ymax=0.3,
ylabel style={font=\color{white!15!black}},
ylabel={probability},
axis background/.style={fill=white},
axis x line*=bottom,
axis y line*=left,
legend style={legend cell align=left, align=left, draw=white!15!black}
]
\addplot[ybar stacked, fill=mycolor1, draw=black, area legend] table[row sep=crcr] {%
0	0.00286815176670399\\
5	0.00105495237396009\\
10	0\\
15	0\\
20	0\\
25	0\\
30	0\\
35	0\\
40	0\\
45	0\\
50	0\\
55	0\\
60	0\\
65	0\\
70	0\\
75	0\\
80	0\\
85	0\\
90	0\\
95	0\\
100	0\\
105	0\\
110	0\\
115	0\\
120	0\\
125	0\\
130	0\\
135	0\\
140	0\\
145	0\\
150	0\\
155	0\\
160	0\\
165	0\\
170	0\\
175	0\\
180	0\\
185	0\\
190	0\\
195	0\\
200	0\\
205	0\\
210	0\\
215	0\\
220	0\\
225	0\\
230	0\\
235	0\\
240	0\\
245	0\\
250	0\\
255	0\\
260	0\\
265	0\\
270	0\\
275	0\\
280	0\\
285	0\\
290	0\\
295	0\\
300	0\\
305	0\\
310	0\\
315	0\\
320	0\\
325	0\\
330	0\\
335	0\\
340	0\\
345	0\\
350	0\\
355	0\\
360	0\\
365	0\\
370	0\\
375	0\\
380	0\\
385	0\\
390	0\\
395	0\\
400	0\\
405	0\\
};
\addplot[forget plot, color=white!15!black] table[row sep=crcr] {%
-2.5	0\\
50	0\\
};
\addlegendentry{$\tau\text{ \textless{}0.5}$}

\addplot[ybar stacked, fill=mycolor2, draw=black, area legend] table[row sep=crcr] {%
0	0.0260182338836719\\
5	0.0144679182714526\\
10	0.00169181302367793\\
15	0\\
20	0.000154253540394164\\
25	0\\
30	0\\
35	0\\
40	0\\
45	0\\
50	0\\
55	0\\
60	0\\
65	0\\
70	0\\
75	0\\
80	0\\
85	0\\
90	0\\
95	0\\
100	0\\
105	0\\
110	0\\
115	0\\
120	0\\
125	0\\
130	0\\
135	0\\
140	0\\
145	0\\
150	0\\
155	0\\
160	0\\
165	0\\
170	0\\
175	0\\
180	0\\
185	0\\
190	0\\
195	0\\
200	0\\
205	0\\
210	0\\
215	0\\
220	0\\
225	0\\
230	0\\
235	0\\
240	0\\
245	0\\
250	0\\
255	0\\
260	0\\
265	0\\
270	0\\
275	0\\
280	0\\
285	0\\
290	0\\
295	0\\
300	0\\
305	0\\
310	0\\
315	0\\
320	0\\
325	0\\
330	0\\
335	0\\
340	0\\
345	0\\
350	0\\
355	0\\
360	0\\
365	0\\
370	0\\
375	0\\
380	0\\
385	0\\
390	0\\
395	0\\
400	0\\
405	0\\
};
\addplot[forget plot, color=white!15!black] table[row sep=crcr] {%
-2.5	0\\
50	0\\
};
\addlegendentry{$\text{0.5\textless}\tau\text{\textless{}0.707}$}

\addplot[ybar stacked, fill=mycolor3, draw=black, area legend] table[row sep=crcr] {%
0	0.0700648503009117\\
5	0.041444557548432\\
10	0.0129705665148641\\
15	0.00167381501278774\\
20	0.00077126770197082\\
25	0.000378217815389537\\
30	0\\
35	9.64084627463525e-05\\
40	9.50112386485793e-05\\
45	9.78473950261488e-05\\
50	0.000168096806839794\\
55	9.50112386485793e-05\\
60	8.29844995791389e-05\\
65	7.66757364532394e-05\\
70	0\\
75	0.000337926570451133\\
80	0.000334478340140407\\
85	6.09839578302509e-05\\
90	0.000130463193368198\\
95	6.75853140902265e-05\\
100	0\\
105	0\\
110	0\\
115	0.000114511361864663\\
120	0\\
125	0\\
130	6.21400518175542e-05\\
135	0\\
140	0\\
145	5.41799625351403e-05\\
150	0.000127709911689973\\
155	0\\
160	5.08199648585424e-05\\
165	0\\
170	8.9805143380164e-05\\
175	9.89551013849354e-05\\
180	4.5056876060151e-05\\
185	5.18243119901342e-05\\
190	0\\
195	6.27346934617413e-05\\
200	0\\
205	0\\
210	5.75068023399296e-05\\
215	0\\
220	0\\
225	0\\
230	0.000184669731457802\\
235	5.70067431891476e-05\\
240	0\\
245	0\\
250	0\\
255	0\\
260	0\\
265	0\\
270	0\\
275	0\\
280	0.000139484584398978\\
285	0\\
290	0\\
295	0\\
300	0.000101639929717085\\
305	0\\
310	0.000103240558531527\\
315	0\\
320	0\\
325	0\\
330	0.000107471728963147\\
335	0\\
340	0.000129817335975287\\
345	0\\
350	0\\
355	0\\
360	0.000128544616995137\\
365	0.000177183120723026\\
370	0\\
375	0\\
380	0\\
385	0\\
390	0\\
395	0\\
400	0\\
405	0\\
};
\addplot[forget plot, color=white!15!black] table[row sep=crcr] {%
-2.5	0\\
50	0\\
};
\addlegendentry{$\text{0.707\textless}\tau\text{\textless{}0.866}$}

\addplot[ybar stacked, fill=mycolor4, draw=black, area legend] table[row sep=crcr] {%
0	0.0473245041506158\\
5	0.0333063535207399\\
10	0.012265644421665\\
15	0.00292917627237854\\
20	0.00200529602512413\\
25	0.00113465344616861\\
30	0.00143235430366009\\
35	0.000867676164717173\\
40	0.000570067431891476\\
45	0.000587084370156893\\
50	0.000420242017099485\\
55	0.000475056193242896\\
60	0.000414922497895694\\
65	0.000306702945812958\\
70	0.000390224730163808\\
75	0.000743438454992492\\
80	0.000334478340140407\\
85	0.000182951873490753\\
90	0.000587084370156893\\
95	0.000540682512721812\\
100	0.000490806719435976\\
105	0.000536730155172676\\
110	0.000258355683418797\\
115	0.000229022723729326\\
120	0.000287534011699648\\
125	0.000401374008168488\\
130	6.21400518175542e-05\\
135	0.000281363753937853\\
140	0.000200176350129831\\
145	0.000270899812675701\\
150	0.000255419823379947\\
155	0.000292450950189679\\
160	0.000254099824292712\\
165	0.000264345784949676\\
170	0.00022451285845041\\
175	0.000445297956232209\\
180	0.000315398132421057\\
185	0.000310945871940805\\
190	5.57938337595912e-05\\
195	0.000125469386923483\\
200	0.000159896962603707\\
205	0.000303508123460739\\
210	5.75068023399296e-05\\
215	0.000182104874076444\\
220	0.000352777155161541\\
225	0.000126681651531439\\
230	0.000307782885763003\\
235	0.000399047202324033\\
240	0.000439142854232189\\
245	0.000139484584398978\\
250	0.000266224384436628\\
255	0.000227367935263074\\
260	0.00020275594227068\\
265	0.000193766762564098\\
270	0.0002327494248551\\
275	0.000238391835154617\\
280	0.000139484584398978\\
285	0.000323741998358122\\
290	8.62602035098944e-05\\
295	0.000354366241446052\\
300	0.000304919789151254\\
305	0\\
310	0.000516202792657635\\
315	0.000206481117063054\\
320	0.00010573831397987\\
325	0.000485612997537183\\
330	0.000429886915852588\\
335	0.000115013604679859\\
340	0.000129817335975287\\
345	0.000138016325615831\\
350	0.000308507080788328\\
355	0.000121403249384296\\
360	0\\
365	0.000177183120723026\\
370	0.000570067431891476\\
375	0\\
380	0\\
385	0\\
390	0\\
395	0.000262231018670079\\
400	0\\
405	0\\
};
\addplot[forget plot, color=white!15!black] table[row sep=crcr] {%
-2.5	0\\
50	0\\
};
\addlegendentry{$\text{0.866\textless}\tau\text{\textless{}0.922}$}

\addplot[ybar stacked, fill=mycolor5, draw=black, area legend] table[row sep=crcr] {%
0	0.0678113024842157\\
5	0.0521447887700272\\
10	0.0267870395415672\\
15	0.0124141280115091\\
20	0.00709566285813154\\
25	0.0044125411795446\\
30	0.00396651961013565\\
35	0.00530246545104939\\
40	0.00408548326188891\\
45	0.00313111664083676\\
50	0.0027735973128566\\
55	0.00370543830729459\\
60	0.00307042648442814\\
65	0.00283700224876986\\
70	0.00312179784131046\\
75	0.00236548599315793\\
80	0.00187307870478628\\
85	0.00182951873490753\\
90	0.00208741109389117\\
95	0.00182480348043612\\
100	0.00175288114084277\\
105	0.00184021767487775\\
110	0.00187307870478628\\
115	0.00137413634237596\\
120	0.00264531290763676\\
125	0.00167239170070203\\
130	0.0014913612436213\\
135	0.00151936427126441\\
140	0.00230202802649306\\
145	0.00140867902591365\\
150	0.00114938920520976\\
155	0.00204715665132775\\
160	0.00254099824292712\\
165	0.0019561588086276\\
170	0.00157159000915287\\
175	0.0016822367235439\\
180	0.00162204753816544\\
185	0.0022802697275659\\
190	0.00156222734526855\\
195	0.00156836733654353\\
200	0.00106597975069138\\
205	0.00176034711607229\\
210	0.00138016325615831\\
215	0.00248876661237806\\
220	0.00164629339075386\\
225	0.00133015734108011\\
230	0.00141580127450982\\
235	0.00182421578205272\\
240	0.00175657141692876\\
245	0.00174355730498723\\
250	0.00153079021051061\\
255	0.00197052210561331\\
260	0.00155446222407521\\
265	0.000904244891965789\\
270	0.0013964965491306\\
275	0.0018276707361854\\
280	0.0012553612595908\\
285	0.00145683899261155\\
290	0.00189772447721768\\
295	0.00168323964686875\\
300	0.00223607845377587\\
305	0.00188655409115165\\
310	0.00216805172916207\\
315	0.0011356461438468\\
320	0.00232624290755715\\
325	0.00157824224199584\\
330	0.00225690630822609\\
335	0.00287534011699648\\
340	0.00233671204755516\\
345	0.0017942122330058\\
350	0.0016967889443358\\
355	0.00121403249384296\\
360	0.000899812318965957\\
365	0.00194901432795329\\
370	0.00171020229567443\\
375	0.00102433991668\\
380	0.000702404514294854\\
385	0.00238391835154617\\
390	0.00131115509335039\\
395	0.000786693056010236\\
400	0\\
405	0.00196673264002559\\
};
\addplot[forget plot, color=white!15!black] table[row sep=crcr] {%
-2.5	0\\
50	0\\
};
\addlegendentry{$\text{0.922\textless}\tau\text{\textless{}0.975}$}

\addplot[ybar stacked, fill=mycolor6, draw=black, area legend] table[row sep=crcr] {%
0	0.0565435634007357\\
5	0.0364712106426202\\
10	0.022839475819652\\
15	0.0117167050895142\\
20	0.00802118410049653\\
25	0.00617755765136243\\
30	0.00815340142083438\\
35	0.00539887391379574\\
40	0.00579568555756334\\
45	0.00567514891151663\\
50	0.00470671059151424\\
55	0.00418049450053749\\
60	0.00456414747685264\\
65	0.00245362356650366\\
70	0.00390224730163808\\
75	0.00324409507633087\\
80	0.00307720072929174\\
85	0.00317116580717305\\
90	0.00450098017120285\\
95	0.00297375381996997\\
100	0.00273449457971473\\
105	0.00429384124138141\\
110	0.00277732359675207\\
115	0.00349259653687223\\
120	0.00327788773337598\\
125	0.00314409639731982\\
130	0.00354198295360059\\
135	0.00348891054882937\\
140	0.00345304203973959\\
145	0.00292571797689757\\
150	0.00285218802774274\\
155	0.00282702585183356\\
160	0.00233771838349295\\
165	0.00296067279143637\\
170	0.00282886201647517\\
175	0.00282022038947066\\
180	0.00225284380300755\\
185	0.00259121559950671\\
190	0.00161802117902815\\
195	0.00225844896462269\\
200	0.0023984544390556\\
205	0.00285297636053095\\
210	0.00235777889593711\\
215	0.00254946823707021\\
220	0.0019990705459154\\
225	0.00247029220486306\\
230	0.00264693281756183\\
235	0.00159618880929613\\
240	0.00276032651231662\\
245	0.00223175335038365\\
250	0.00212979507549303\\
255	0.00227367935263074\\
260	0.00270341256360906\\
265	0.00251896791333327\\
270	0.00271540995664283\\
275	0.00262231018670079\\
280	0.00258046481138109\\
285	0.00339929098276028\\
290	0.00224276529125725\\
295	0.00283492993156842\\
300	0.00396395725896631\\
305	0.00377310818230329\\
310	0.00258101396328818\\
315	0.00402638178272956\\
320	0.00496970075705391\\
325	0.00315648448399169\\
330	0.00429886915852588\\
335	0.00356542174507563\\
340	0.00298579872743159\\
345	0.00317437548916411\\
350	0.00262231018670079\\
355	0.00339929098276028\\
360	0.00231380310591246\\
365	0.00318929617301447\\
370	0.00304035963675454\\
375	0.00307301975003999\\
380	0.00374615740957255\\
385	0.00238391835154617\\
390	0.00183561713069055\\
395	0.00340900324271102\\
400	0.00124871913652418\\
405	0.00655577546675197\\
};
\addplot[forget plot, color=white!15!black] table[row sep=crcr] {%
-2.5	0\\
50	0\\
};
\addlegendentry{$\text{0.975\textless}\tau\text{\textless{}1.0}$}

\end{axis}

\begin{axis}[%
width=2.691in,
height=1.254in,
at={(0.514in,0.532in)},
scale only axis,
xmin=0,
xmax=400,
xlabel style={font=\color{white!15!black}},
xlabel={distance, m},
ymin=0,
ymax=1,
ylabel style={font=\color{white!15!black}},
ylabel={cumulative density},
axis background/.style={fill=white},
axis x line*=bottom,
axis y line*=left,
legend style={at={(0.97,0.03)}, anchor=south east, legend cell align=left, align=left, draw=white!15!black}
]
\addplot [color=red, draw=none, mark=o, mark options={solid, red}]
  table[row sep=crcr]{%
0	0\\
5	0.391160563507268\\
10	0.666454140172779\\
15	0.767835791599883\\
20	0.792046890308263\\
25	0.808621659283972\\
30	0.818000134333097\\
35	0.829839255030033\\
40	0.837011030067599\\
45	0.841722920913536\\
50	0.846575465217561\\
55	0.850048974495015\\
60	0.853975550199963\\
65	0.85740509100555\\
70	0.859940143390499\\
75	0.86316554486242\\
80	0.869310433439854\\
85	0.87207506327293\\
90	0.873587251497905\\
95	0.87843979580193\\
100	0.882908805676427\\
105	0.886965567313839\\
110	0.891401908987499\\
115	0.893537347203392\\
120	0.89543033391879\\
125	0.89780694552968\\
130	0.90112450132937\\
135	0.901638119762814\\
140	0.903963731124542\\
145	0.905618288215482\\
150	0.907857409898469\\
155	0.909968581770999\\
160	0.912385834323934\\
165	0.91448609574751\\
170	0.916671045131715\\
175	0.918526755567615\\
180	0.922207364643498\\
185	0.924814287070288\\
190	0.927384409349891\\
195	0.927845573134812\\
200	0.928882640019564\\
205	0.930204267939766\\
210	0.932712913529038\\
215	0.933188235851216\\
220	0.934693423204779\\
225	0.937609301844866\\
230	0.938656388699519\\
235	0.94120036732526\\
240	0.944498690917416\\
245	0.948128425014047\\
250	0.949281334476351\\
255	0.951481811419631\\
260	0.953361120485352\\
265	0.955036999188288\\
270	0.956638577850208\\
275	0.958562367603875\\
280	0.960532794684903\\
285	0.961685704147207\\
290	0.964361592775764\\
295	0.965074576259031\\
300	0.968003589487586\\
305	0.970523903195878\\
310	0.970523903195878\\
315	0.974790576009129\\
320	0.976497245134429\\
325	0.977371224888111\\
330	0.981385057830946\\
335	0.984938286993457\\
340	0.985888931637812\\
345	0.986961936483917\\
350	0.988102710057144\\
355	0.990652674514945\\
360	0.991656132750654\\
365	0.991656132750654\\
370	0.993120639364932\\
375	0.997832530210869\\
380	0.997832530210869\\
385	0.997832530210869\\
390	0.997832530210869\\
395	0.997832530210869\\
400	1\\
405	1\\
410	1\\
};
\addlegendentry{$\text{0.866\textless}\tau\text{\textless{}0.922}$}

\addplot [color=black]
  table[row sep=crcr]{%
0	0\\
5	0.381279773850502\\
10	0.650947969971226\\
15	0.786229751718887\\
20	0.835021247559831\\
25	0.848569041088226\\
30	0.852534093448397\\
35	0.854720483862945\\
40	0.856678872176544\\
45	0.858616945071151\\
50	0.860553755050542\\
55	0.862490510135043\\
60	0.864427263549056\\
65	0.866364016927456\\
70	0.868300770305325\\
75	0.870237523683188\\
80	0.872174277061051\\
85	0.874111030438913\\
90	0.876047783816776\\
95	0.877984537194639\\
100	0.879921290572502\\
105	0.881858043950365\\
110	0.883794797328228\\
115	0.885731550706091\\
120	0.887668304083954\\
125	0.889605057461816\\
130	0.891541810839679\\
135	0.893478564217542\\
140	0.895415317595405\\
145	0.897352070973268\\
150	0.899288824351131\\
155	0.901225577728994\\
160	0.903162331106857\\
165	0.905099084484719\\
170	0.907035837862582\\
175	0.908972591240445\\
180	0.910909344618308\\
185	0.912846097996171\\
190	0.914782851374034\\
195	0.916719604751897\\
200	0.91865635812976\\
205	0.920593111507622\\
210	0.922529864885485\\
215	0.924466618263348\\
220	0.926403371641211\\
225	0.928340125019074\\
230	0.930276878396937\\
235	0.9322136317748\\
240	0.934150385152662\\
245	0.936087138530525\\
250	0.938023891908388\\
255	0.939960645286251\\
260	0.941897398664114\\
265	0.943834152041977\\
270	0.94577090541984\\
275	0.947707658797703\\
280	0.949644412175565\\
285	0.951581165553428\\
290	0.953517918931291\\
295	0.955454672309154\\
300	0.957391425687017\\
305	0.95932817906488\\
310	0.961264932442743\\
315	0.963201685820606\\
320	0.965138439198468\\
325	0.967075192576331\\
330	0.969011945954194\\
335	0.970948699332057\\
340	0.97288545270992\\
345	0.974822206087783\\
350	0.976758959465646\\
355	0.978695712843508\\
360	0.980632466221371\\
365	0.982569219599234\\
370	0.984505972977097\\
375	0.98644272635496\\
380	0.988379479732823\\
385	0.990316233110686\\
390	0.992252986488549\\
395	0.994189739866411\\
400	0.996126493244274\\
405	0.998063246622137\\
410	1\\
};
\addlegendentry{fitted distribution}

\addplot [color=green, draw=none, mark=o, mark options={solid, green}]
  table[row sep=crcr]{%
0	0\\
5	0.146774526541186\\
10	0.241445685422991\\
15	0.300731889692081\\
20	0.331145852638636\\
25	0.351967064246866\\
30	0.368002631249136\\
35	0.389167049642117\\
40	0.403181326686117\\
45	0.418225638585319\\
50	0.43295706413477\\
55	0.445174638993642\\
60	0.456026273806181\\
65	0.467873786497086\\
70	0.474242848077651\\
75	0.484372214988706\\
80	0.492793164965071\\
85	0.500780894300646\\
90	0.509012536099249\\
95	0.520696080500537\\
100	0.528415284645539\\
105	0.535513424966526\\
110	0.546659282732515\\
115	0.553868597665073\\
120	0.562934602215683\\
125	0.57144327042097\\
130	0.579604646046448\\
135	0.588798846761165\\
140	0.597855283305847\\
145	0.606818613170958\\
150	0.614413126940965\\
155	0.621816773301409\\
160	0.629155104244258\\
165	0.635223301723997\\
170	0.642908550425546\\
175	0.650251647643807\\
180	0.657572313118543\\
185	0.663420195046575\\
190	0.670146414971702\\
195	0.674346438426227\\
200	0.680208870108338\\
205	0.686434724892694\\
210	0.693840417589887\\
215	0.699960687702461\\
220	0.706578540751017\\
225	0.711767683692358\\
230	0.718180013354312\\
235	0.725050862797079\\
240	0.729194214270957\\
245	0.736359408549093\\
250	0.742152544348437\\
255	0.747681019121414\\
260	0.753582985506584\\
265	0.760600443820222\\
270	0.767139124805565\\
275	0.774187725685684\\
280	0.780994660249913\\
285	0.787692973517904\\
290	0.796516777582645\\
295	0.80233849793363\\
300	0.809697346111175\\
305	0.819986898359428\\
310	0.829781048811556\\
315	0.836480787555876\\
320	0.846932379997015\\
325	0.8598326188889\\
330	0.868026151234731\\
335	0.879185060356418\\
340	0.888440102965677\\
345	0.896190573014056\\
350	0.904430546433912\\
355	0.911237480998141\\
360	0.920061285062882\\
365	0.926067403796025\\
370	0.934346107995763\\
375	0.942238206041246\\
380	0.950215082483702\\
385	0.959939274718315\\
390	0.966127397049432\\
395	0.970892251244392\\
400	0.97974126617789\\
405	0.982982663589427\\
410	1\\
};
\addlegendentry{$\text{0.975\textless}\tau\text{\textless{}1.0}$}

\addplot [color=black, forget plot]
  table[row sep=crcr]{%
0	0\\
5	0.116549247419029\\
10	0.21837259854671\\
15	0.296453779829357\\
20	0.349668112288632\\
25	0.382830292554616\\
30	0.402862802235734\\
35	0.415747301966848\\
40	0.425355134536672\\
45	0.433688715004441\\
50	0.441599872517154\\
55	0.449391265364866\\
60	0.45715355086874\\
65	0.464909762001554\\
70	0.47266488327005\\
75	0.48041983625822\\
80	0.488174766869116\\
85	0.49592969491584\\
90	0.503684622709254\\
95	0.511439550481086\\
100	0.519194478251333\\
105	0.526949406021478\\
110	0.534704333791618\\
115	0.542459261561758\\
120	0.550214189331897\\
125	0.557969117102037\\
130	0.565724044872177\\
135	0.573478972642317\\
140	0.581233900412456\\
145	0.588988828182596\\
150	0.596743755952736\\
155	0.604498683722875\\
160	0.612253611493015\\
165	0.620008539263155\\
170	0.627763467033294\\
175	0.635518394803434\\
180	0.643273322573574\\
185	0.651028250343713\\
190	0.658783178113853\\
195	0.666538105883993\\
200	0.674293033654133\\
205	0.682047961424272\\
210	0.689802889194412\\
215	0.697557816964552\\
220	0.705312744734691\\
225	0.713067672504831\\
230	0.720822600274971\\
235	0.728577528045111\\
240	0.73633245581525\\
245	0.74408738358539\\
250	0.75184231135553\\
255	0.759597239125669\\
260	0.767352166895809\\
265	0.775107094665949\\
270	0.782862022436088\\
275	0.790616950206228\\
280	0.798371877976368\\
285	0.806126805746507\\
290	0.813881733516647\\
295	0.821636661286787\\
300	0.829391589056927\\
305	0.837146516827066\\
310	0.844901444597206\\
315	0.852656372367346\\
320	0.860411300137485\\
325	0.868166227907625\\
330	0.875921155677765\\
335	0.883676083447904\\
340	0.891431011218044\\
345	0.899185938988184\\
350	0.906940866758324\\
355	0.914695794528463\\
360	0.922450722298603\\
365	0.930205650068743\\
370	0.937960577838882\\
375	0.945715505609022\\
380	0.953470433379162\\
385	0.961225361149301\\
390	0.968980288919441\\
395	0.976735216689581\\
400	0.984490144459721\\
405	0.99224507222986\\
410	1\\
};
\end{axis}
\end{tikzpicture}%

%% file: auc_gator.tex
%
%
\definecolor{mycolor1}{rgb}{0.00000,0.44700,0.74100}%
\definecolor{mycolor2}{rgb}{0.85000,0.32500,0.09800}%
\definecolor{mycolor3}{rgb}{0.92900,0.69400,0.12500}%
\definecolor{mycolor4}{rgb}{0.49400,0.18400,0.55600}%
\definecolor{mycolor5}{rgb}{0.46600,0.67400,0.18800}%
\definecolor{mycolor6}{rgb}{0.30100,0.74500,0.93300}%
\definecolor{mycolor7}{rgb}{0.63500,0.07800,0.18400}%
\begin{tikzpicture}

\begin{axis}[%
width=2.826in,
height=1.58in,
at={(0.474in,0.318in)},
scale only axis,
xmin=0,
xmax=30,
xlabel style={font=\color{white!15!black}},
xlabel={Recall \%},
ymin=0,
ymax=100,
ylabel style={font=\color{white!15!black}},
ylabel={Precision \%},
axis background/.style={fill=white},
axis x line*=bottom,
axis y line*=left,
xmajorgrids,
ymajorgrids,
legend style={at={(1,1)},anchor=north east, draw=white!15!black,font=\fontsize{6}{1}\selectfont}
]
\addplot [color=mycolor1, mark=asterisk, mark options={solid, mycolor1}]
  table[row sep=crcr]{%
0.00564748404585757	100\\
0.0960072287795787	60.7142857142857\\
0.796295024566556	43.3846030769231\\
1.3610430790083	32.6558127371274\\
2.37760112384932	11.3447003772568\\
3.64827191788558	3.64827191788558\\
};
\addlegendentry{USC}

\addplot [color=mycolor2, mark=asterisk, mark options={solid, mycolor2}]
  table[row sep=crcr]{%
0.0620907089636487	36.6666333333333\\
0.756377483630616	33.0048854679803\\
3.01422125197562	17.5599946399211\\
4.56084540584782	12.7324199826662\\
6.78485155791375	9.37893494069912\\
8.19604649017837	8.19604649017837\\
8.19604649017837	0\\
};
\addlegendentry{FPFH}

\addplot [color=mycolor3, mark=asterisk, mark options={solid, mycolor3}]
  table[row sep=crcr]{%
0.197650666365485	56.4515806451613\\
1.11813863790377	54.5454517906336\\
3.59724538061893	38.3966372513562\\
5.33091474474814	27.0874715351507\\
7.87778869155184	14.6149693190152\\
9.57753901626384	9.57753901626384\\
9.57753901626384	0\\
};
\addlegendentry{Gestalt3D}

\addplot [color=mycolor4, mark=asterisk, mark options={solid, mycolor4}]
  table[row sep=crcr]{%
0	0\\
0.163693723188079	74.3589230769231\\
2.56265584782118	67.0605775480059\\
5.26642883269361	58.8643868769716\\
9.9514720365771	35.4943182202537\\
13.1520318299842	14.649569060044\\
13.1520318299842	0\\
};
\addlegendentry{NBLD}

\addplot [color=mycolor5, mark=asterisk, mark options={solid, mycolor5}]
  table[row sep=crcr]{%
0.118536972228494	77.7778148148148\\
1.77805447053511	79.7468683544304\\
7.40574112102055	61.3084624766355\\
11.1311962914879	45.8392081589958\\
16.0645513772861	27.4312859951807\\
19.0448908726575	19.0448908726575\\
19.0448908726575	0\\
};
\addlegendentry{SHOT}

\addplot [color=mycolor6, mark=asterisk, mark options={solid, mycolor6}]
  table[row sep=crcr]{%
0.191916911266652	94.4444444444444\\
2.74328234364416	92.9254110898661\\
10.2844889139761	77.3673059872611\\
14.8453310002258	60.5990516129032\\
21.3987407936329	36.6989440367861\\
25.2765969180402	25.2765969180402\\
25.2765969180402	0\\
};
\addlegendentry{ISHOT}

\addplot [color=mycolor7, mark=asterisk, mark options={solid, mycolor7}]
  table[row sep=crcr]{%
0.101648972215948	94.7368421052632\\
2.26451310142309	92.6096951501155\\
9.49853026880506	78.4149062937063\\
14.6092326067314	62.1576864488227\\
21.6738334876892	38.1359542329094\\
25.9996371809352	25.9996371809352\\
25.9996371809352	0\\
};
\addlegendentry{ISHOT-C}

\end{axis}
\end{tikzpicture}%

%% file: auc_pole.tex
%
%
\definecolor{mycolor1}{rgb}{0.00000,0.44700,0.74100}%
\definecolor{mycolor2}{rgb}{0.85000,0.32500,0.09800}%
\definecolor{mycolor3}{rgb}{0.92900,0.69400,0.12500}%
\definecolor{mycolor4}{rgb}{0.49400,0.18400,0.55600}%
\definecolor{mycolor5}{rgb}{0.46600,0.67400,0.18800}%
\definecolor{mycolor6}{rgb}{0.30100,0.74500,0.93300}%
\definecolor{mycolor7}{rgb}{0.63500,0.07800,0.18400}%
\begin{tikzpicture}

\begin{axis}[%
width=2.826in,
height=1.58in,
at={(0.474in,0.318in)},
scale only axis,
xmin=0,
xmax=20,
xlabel style={font=\color{white!15!black}},
xlabel={Recall \%},
ymin=0,
ymax=100,
ylabel style={font=\color{white!15!black}},
ylabel={Precision \%},
axis background/.style={fill=white},
axis x line*=bottom,
axis y line*=left,
xmajorgrids,
ymajorgrids,
legend style={at={(1,1)},anchor=north east, draw=white!15!black,font=\fontsize{6}{1}\selectfont}
]
\addplot [color=mycolor1, mark=asterisk, mark options={solid, mycolor1}]
  table[row sep=crcr]{%
0	0\\
0.0871011164779476	61.1111111111111\\
0.65721719059308	39.52379\\
1.21149544698709	28.9772272727273\\
2.17753038086943	10.7927908869702\\
3.4286153131681	3.4286153131681\\
};
\addlegendentry{USC}

\addplot [color=mycolor2, mark=asterisk, mark options={solid, mycolor2}]
  table[row sep=crcr]{%
0.110768177862173	55.99996\\
0.941529677980853	35.1032259587021\\
3.37844104755123	19.0540456938867\\
4.80258054909408	13.4589391485588\\
6.94677638737242	9.63039451135242\\
8.47377366801171	8.47377366801171\\
8.47377366801171	0\\
};
\addlegendentry{FPFH}

\addplot [color=mycolor3, mark=asterisk, mark options={solid, mycolor3}]
  table[row sep=crcr]{%
0.0237360550676478	21.4285714285714\\
0.269008386739457	25.3731119402985\\
1.29757137431759	18.7858013745705\\
2.35777661998576	13.9643574039363\\
4.14590417754569	7.97080664739884\\
5.53840946277395	5.53840946277395\\
5.53840946277395	0\\
};
\addlegendentry{Gestalt3D}

\addplot [color=mycolor4, mark=asterisk, mark options={solid, mycolor4}]
  table[row sep=crcr]{%
0	0\\
0.0237360550676478	37.5\\
0.965264657014004	57.2769014084507\\
2.21536646886621	41.4201431952663\\
4.73929472268376	19.7624368195315\\
6.91509730041934	7.76750042481336\\
6.91509730041934	0\\
};
\addlegendentry{NBLD}

\addplot [color=mycolor5, mark=asterisk, mark options={solid, mycolor5}]
  table[row sep=crcr]{%
0.17406424558905	73.3332666666667\\
1.40042779492048	67.5572782442748\\
5.80742457472901	48.2577509533202\\
9.04343353113379	36.8353066065098\\
13.6798718965108	23.364851472973\\
16.2592215436348	16.2592215436348\\
16.2592215436348	0\\
};
\addlegendentry{SHOT}

\addplot [color=mycolor6, mark=asterisk, mark options={solid, mycolor6}]
  table[row sep=crcr]{%
0.102856238626474	100\\
1.70108378827439	89.583325\\
6.99423270828388	68.7938577431906\\
10.4438530975552	48.4225455979457\\
15.4759177466572	27.2689424787397\\
18.7435268138302	18.7435268138302\\
18.7435268138302	0\\
};
\addlegendentry{ISHOT}

\addplot [color=mycolor7, mark=asterisk, mark options={solid, mycolor7}]
  table[row sep=crcr]{%
0.0553841284911781	100\\
1.33713070654324	88.9473421052631\\
6.5986193527969	71.5879399141631\\
10.3172647677823	51.0371465362035\\
15.8873166627107	28.6570280148423\\
19.2815991771501	19.2815991771501\\
19.2815991771501	0\\
};
\addlegendentry{ISHOT-C}

\end{axis}
\end{tikzpicture}%